\documentclass[10pt,twocolumn,letterpaper]{article}

\usepackage{cvpr}              

\usepackage{graphicx}
\usepackage{booktabs}
\usepackage{multirow,multicol,makecell}
\usepackage[dvipsnames]{xcolor}
\usepackage[normalem]{ulem}
\usepackage{enumitem}
\usepackage{comment}
\usepackage{tabularray}
\usepackage{amsmath,amssymb,amsthm,amsfonts,mathrsfs,bm,color}
\usepackage{caption}
\usepackage{dsfont}
\usepackage{wrapfig}

\usepackage[pagebackref,breaklinks,colorlinks]{hyperref}

\DeclareMathOperator*{\minimize}{\text{minimize}}

\DeclareMathOperator*{\st}{\text{subject to}}
\DeclareMathAlphabet\mathbfcal{OMS}{cmsy}{b}{n}

\renewcommand\footnotemark{}
\usepackage[capitalize]{cleveref}
\crefname{section}{Sec.}{Secs.}
\Crefname{section}{Section}{Sections}
\Crefname{table}{Table}{Tables}
\crefname{table}{Tab.}{Tabs.}

\def\cvprPaperID{10308} 
\def\confName{CVPR}
\def\confYear{2023}

\begin{document}

\title{Towards High-Quality and Efficient Video Super-Resolution via Spatial-Temporal Data Overfitting}

\author{
Gen Li\text{$^{1,\dagger}$}\thanks{$\dagger$ Equal Contribution.},
Jie Ji\text{$^{1,\dagger}$},
Minghai Qin\text{$^{\dagger}$},
Wei Niu$^2$, 
Bin Ren$^2$, 
Fatemeh Afghah$^1$,
Linke Guo$^1$,
Xiaolong Ma$^1$ \and
\centerline{$^1$Clemson University~$^2$William \& Mary} \and
{\tt\small \{gen,xiaolom\}@clemson.edu}
}
\maketitle

\begin{abstract}

As deep convolutional neural networks (DNNs) are widely used in various fields of computer vision, leveraging the overfitting ability of the DNN to achieve video resolution upscaling has become a new trend in the modern video delivery system. By dividing videos into chunks and overfitting each chunk with a super-resolution model, the server encodes videos before transmitting them to the clients, thus achieving better video quality and transmission efficiency. However, a large number of chunks are expected to ensure good overfitting quality, which substantially increases the storage and consumes more bandwidth resources for data transmission. On the other hand, decreasing the number of chunks through training optimization techniques usually requires high model capacity, which significantly slows down execution speed. To reconcile such, we propose a novel method for high-quality and efficient video resolution upscaling tasks, which leverages the spatial-temporal information to accurately divide video into chunks, thus keeping the number of chunks as well as the model size to minimum. Additionally, we advance our method into a single overfitting model by a data-aware joint training technique, which further reduces the storage requirement with negligible quality drop. We deploy our models on an off-the-shelf mobile phone, and experimental results show that our method achieves real-time video super-resolution with high video quality. Compared with the state-of-the-art, our method achieves 28 fps streaming speed with 41.6 PSNR, which is 14$\times$ faster and 2.29 dB better in the live video resolution upscaling tasks. 
Code available in \url{https://github.com/coulsonlee/STDO-CVPR2023.git}.

\end{abstract}


\section{Introduction}

Being praised by its high image quality performance and wide application scenarios, deep learning-based super-resolution (SR) becomes the core enabler of many incredible, cutting-edge applications in the field of image/video reparation~\cite{dong2015image,dong2016accelerating,shi2016real,sajjadi2018frame}, surveillance system enhancement~\cite{deshmukh2019fractional}, medical image processing~\cite{peng2020saint}, and high-quality video live streaming~\cite{kim2020neural}. 
Distinct from the traditional methods that adopt classic interpolation algorithms~\cite{feichtinger1992iterative,tom1994reconstruction} to improve the image/video quality, the deep learning-based approaches ~\cite{dong2015image,dong2016accelerating,shi2016real,ledig2017photo,kim2016accurate,tong2017image,zhang2018image,lim2017enhanced,timofte2017ntire,yu2018wide} exploit the advantages of learning a mapping function from low-resolution (LR) to high-resolution (HR) using external data, thus achieving better performance due to better generalization ability when meeting new data.

\begin{figure}[t]
\centering
\includegraphics[width=1\columnwidth]{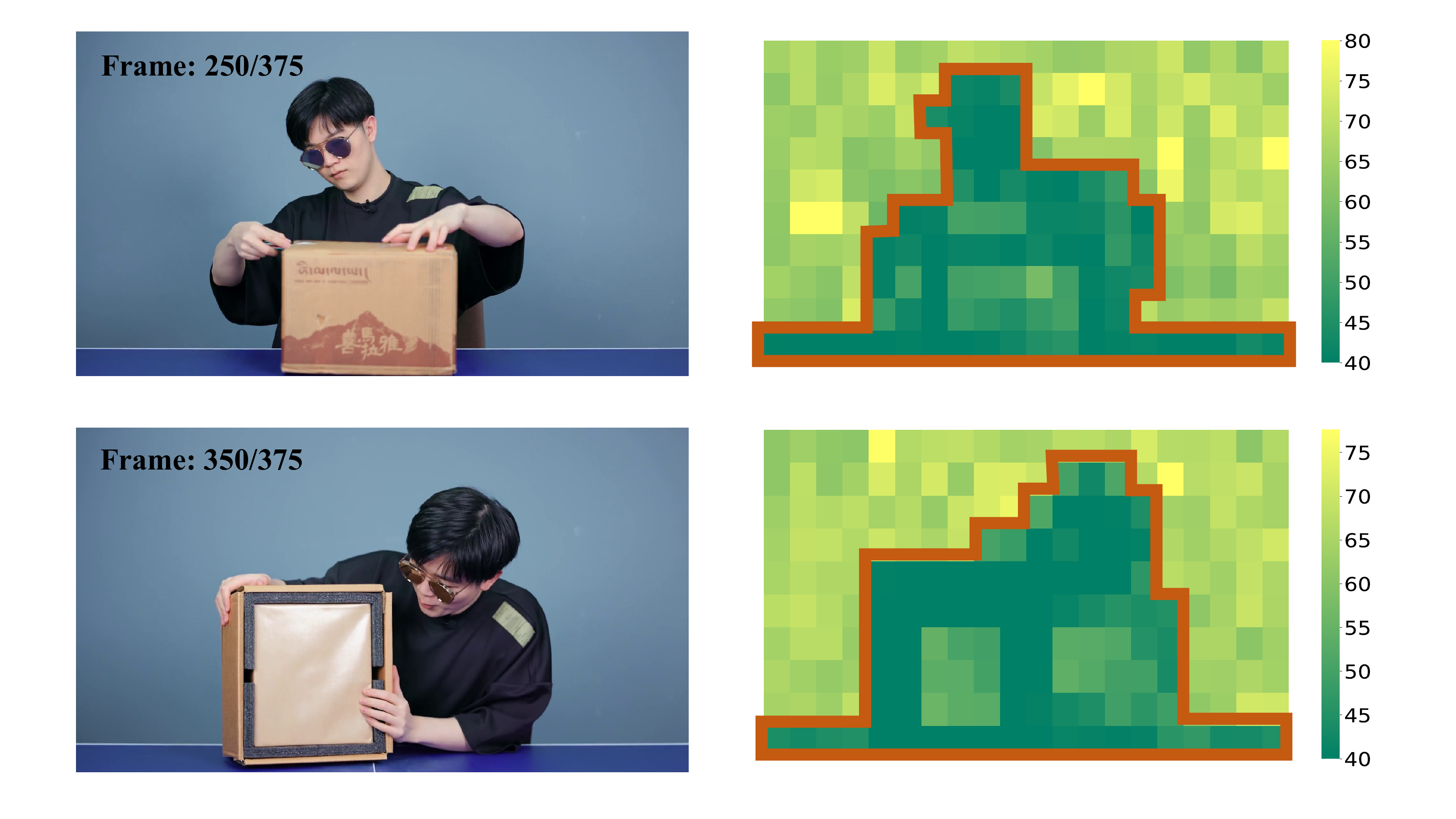}
\vspace{-0.8cm}
\caption{Patch PSNR heatmap of two frames in a 15s video when super-resolved by a general WDSR model. A clear boundary shows that PSNR is strongly related to video content.}
\label{fig:intro_1}
\vspace{-0.6cm}
\end{figure}

\begin{figure*}[t!]
\centering
\includegraphics[width=0.88\textwidth]{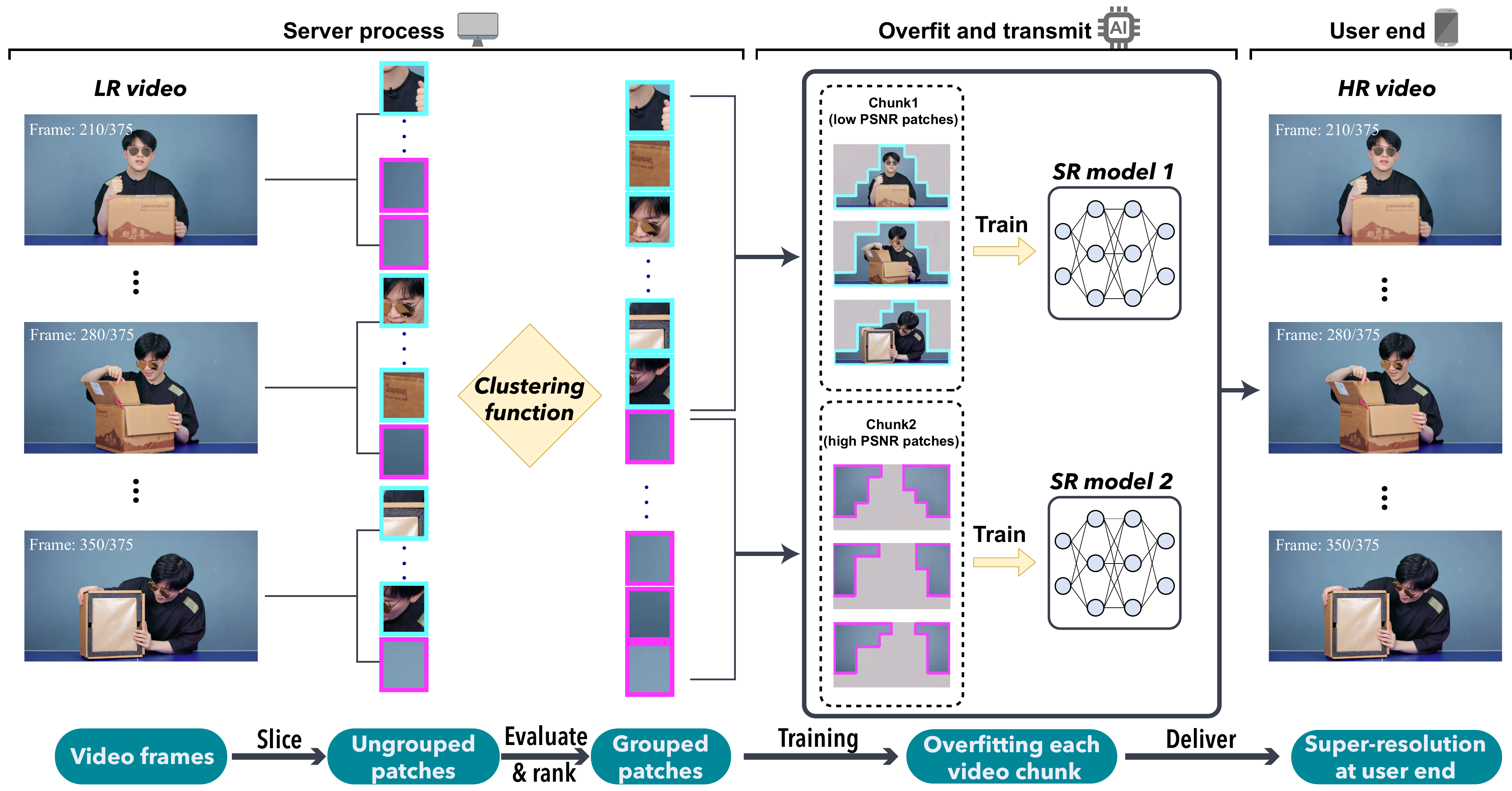}
\vspace{-0.2cm}
\caption{Overview of the proposed STDO method. Each video frame is sliced into patches, and all patches across time dimension are divided and grouped into chunks. Here we set the number of chunks to 2 for clear illustration. Then each chunk is overfitted by independent SR models, and delivered to end-user for video super-resolution.}
\label{fig:intro_2}
\vspace{-0.5cm}
\end{figure*}

Such benefits have driven numerous interests in designing new methods~\cite{wu2018video,chen2017deepcoder,han2018deep} to deliver high-quality video stream to users in the real-time fashion, especially in the context of massive online video and live streaming available. Among this huge family, an emerging representative~\cite{dosovitskiy2015flownet,habibian2019video,lu2019dvc,rippel2019learned} studies the prospect of utilizing \textit{SR model} to upscale the resolution of the LR video in lieu of transmitting the HR video directly, which in many cases, consumes tremendous bandwidth between servers and clients~\cite{khani2021efficient}. One practical method is to deploy a pretrained SR model on the devices of the end users~\cite{yeo2017will,lee2019mobisr}, and perform resolution upscaling for the transmitted LR videos, thus obtaining HR videos without causing bandwidth congestion.
However, the deployed SR model that is trained with limited data usually suffers from limited generalization ability, and may not achieve good performance at the presence of new data distribution~\cite{yeo2018neural}. To overcome this limitation, new approaches~\cite{yeo2018neural,xiao2019sensor,yeo2020nemo,kim2020neural,chen2020sr360,dasari2020streaming,liu2021overfitting} exploit the overfitting property of DNN by training an SR model for each video chunk (i.e., a fragment of the video), and delivering the video alongside the corresponding SR models to the clients. This trade-off between model expressive power and the storage efficiency significantly improves the quality of the resolution upscaled videos. However, to obtain better overfitting quality, more video segments are expected, which notably increase the data volume as well as system overhead when processing the LR videos~\cite{yeo2018neural}. While advanced training techniques are proposed to reduce the number of SR models~\cite{liu2021overfitting}, it still requires overparameterized SR backbones (e.g., EDSR~\cite{lim2017enhanced}) and handcrafted modules to ensure sufficient model capacity for the learning tasks, which degrades the execution speed at user-end when the device is resource-constraint.

In this work, we present a novel approach towards high-quality and efficient video resolution upscaling via \underline{\bf S}patial-\underline{\bf T}emporal \underline{\bf D}ata \underline{\bf O}verfitting, namely \textbf{STDO}, which for the first time, utilizes the spatial-temporal information to accurately divide video into chunks. Inspired by the work proposed in~\cite{bengio2007scaling,kumar2010self,fan2017learning,toneva2018empirical,yuan2021mest} that images may have different levels of intra- and inter-image (i.e., within one image or between different images) information density due to varied texture complexity, we argue that the unbalanced information density within or between frames of the video universally exists, and should be properly managed for data overfitting.
Our preliminary experiment in Figure~\ref{fig:intro_1} shows that the PSNR values at different locations in a video frame forms certain pattern regarding the video content, and exhibits different patterns along the timeline. Specifically, at the server end, each frame of the video is evenly divided into patches, and then we split all the patches into multiple chunks by PSNR regarding all frames. Independent SR models will be used to overfit the video chunks, and then delivered to the clients. Figure~\ref{fig:intro_2} demonstrates the overview of our proposed method. By using spatial-temporal information for data overfitting, we reduce the number of chunks as well as the overfitting models since they are bounded by the nature of the content, which means our method can keep a minimum number of chunks regardless the duration of videos. In addition, since each chunk has similar data patches, we can actually use smaller SR model without handcrafted modules for the overfitting task, which reduces the computation burden for devices of the end-user. Our experimental results demonstrate that our method achieves real-time video resolution upscaling from 270p to 1080p on an off-the-shelf mobile phone with high PSNR.

Note that STDO encodes different video chunks with independent SR models, we further improve it by a \underline{\bf J}oint training technique (\textbf{JSTDO}) that results in one single SR model for all chunks, which further reduces the storage requirement. We design a novel data-aware joint training technique, which trains a single SR model with more data from higher information density chunks and less data from their counterparts. The underlying rationale is consistent with the discovery in~\cite{toneva2018empirical,yuan2021mest}, that more informative data contributes majorly to the model training. We summarize our contributions as follows:
\begin{itemize}[leftmargin=*]
    \item We discover the unbalanced information density within video frames, and it universally exists and constantly changes along the video timeline.
    
    \item By leveraging the unbalanced information density in the video, we propose a spatial-temporal data overfitting method STDO for video resolution upscaling, which achieves outperforming video quality as well as real-time execution speed.
    
    \item We propose an advanced data-aware joint training technique which takes different chunk information density into consideration, and reduces the number of SR models to a single model with negligible quality degradation.
    
    \item We deploy our models on an off-the-shelf mobile phone, and achieve real-time super-resolution performance.
\end{itemize}

\vspace{-1.1em}
\section{Related Works}
\label{sec:related work}

\subsection{Single Image Super Resolution (SISR)}
For SISR tasks, SRCNN~\cite{dong2015image} is the pioneer of applying DNN to image super resolution. Then, followed by FSRCNN~\cite{dong2016accelerating} and ESPCN~\cite{shi2016real}, both of them make progress in efficiency and performance. After this, with the development of deep neural networks, more and more network backbones are used for SISR tasks. For example, VDSR~\cite{kim2016accurate} uses the VGG~\cite{simonyan2014very} network as the backbone and adds residual learning to further improve the effectiveness. Similarly, SRResNet~\cite{ledig2017photo} proposed a SR network using ResNet~\cite{he2016deep} as a backbone. EDSR ~\cite{lim2017enhanced} removes the batch norm in residual blocks by finding that the use of batch norm will ignore absolute differences between image pixels (or features). WDSR~\cite{yu2018wide} finds that ReLU will impede information transfer, so the growth of the the number of filters before ReLU increases the width of the feature map. With the emergence of channel attention mechanisms networks represented by SENet~\cite{qin2021fcanet}, various applications of attention mechanisms poured into the area of image super resolution~\cite{zhang2018image,dai2019second,niu2020single,zhang2021context}. After witnessing the excellent performance of transformer~\cite{dosovitskiy2020image} in the field of computer vision, more and more researchers apply various vision transformer models into image super resolution tasks~\cite{chen2021pre,liang2021swinir,mei2021image}.

\subsection{Video Super Resolution (VSR)}
The VSR methods mainly learn from SISR framework~\cite{liu2022video}. Some aforementioned works like EDSR and WDSR are all present results on VSR.
Some of the current VSR works perform alignment to estimate the motions between images by computing optical flow by DNNs~\cite{sajjadi2018frame,tao2017detail,caballero2017real,kim2018spatio}. 
The Deformable convolution (DConv)~\cite{dai2017deformable} was first used to deal with geometric transformation in vision tasks because the sampling operation of CNN is fixed. TDAN~\cite{tian2020tdan} applies DConv to align the input frames at the feature level, which avoids the two-stage process used in previous optical flow based methods. EDVR~\cite{wang2019edvr} uses their proposed Pyramid, Cascading and Deformable convolutions (PCD) alignment module and the temporal-spatial attention (TSA) fusion module to further improve the robustness of alignment and take account of the visual informativeness of each frame. Other works like DNLN~\cite{wang2019deformable} and D3Dnet~\cite{ying2020deformable} also apply Dconv in their model to achieve a better alignment performance. 
\subsection{Content-Aware DNN}
It is impractical to develop a DNN model to work well on all the video from Internet. NAS~\cite{yeo2018neural} was the first proposed video delivery framework  to consider using DNN models to overfit each video chunk to guarantee reliability and performance. Other livestreaming and video streaming works~\cite{yeo2020nemo,xiao2019sensor,kim2020neural,chen2020sr360,dasari2020streaming} leverage overfitting property to ensure excellent performance at the client end. \cite{kim2020neural} proposes a live video ingest framework, which adds an online learning module to the original NAS~\cite{yeo2018neural} framework to further ensure quality. NEMO~\cite{yeo2020nemo} selects key frames to apply super-resolution. This greatly reduces the amount of computation on the client sides. CaFM~\cite{liu2021overfitting} splits a long video into different chunks by time and design a handcrafted layer along with a joint training technique to reduce the number of SR models and improve performance. EMT~\cite{li2022efficient} proposes to leverage meta-tuning and challenge patches sampling technique to further reduce model size and computation cost. 
SRVC~\cite{khani2021efficient} encodes video into content and time-varying model streams. Our work differentiates from these works by taking spatial information as well as temporal information into account, which exhibits better training effects for the overfitting tasks.

\section{Proposed Method}

\subsection{Motivation}

To tackle the limited generalization ability caused by using only one general SR model to super-resolve various videos, previous works~\cite{yeo2018neural,kim2020neural,liu2021overfitting} split the video by time and train separate SR models to overfit each of the video chunks. 
With more fine-grained video chunks over time, the better overfitting quality can be obtained, which makes these approaches essentially a trade-off between model expressive power and the storage efficiency. For a specific video, more chunks will surely improve the overfitting quality, but it also inevitably increases the data volume as well as system overhead when performing SR inference.

In the aforementioned methods, images are stacked according to the timeline to form the video. However, patches have spatial location information~\cite{rao2021dynamicvit}, and these patches are fed into the neural network indiscriminately for training, which may cause redundancy that contradicts with overfitting property.
As illustrated in Figure~\ref{fig:intro_1}, when using a general SR model to super-resolve an LR video, the values of PSNR at different patch locations form a clear boundary, and are strongly related to the content of the current video frame (i.e., spatial information). Meanwhile, diverse boundary patterns can be seen in different frames (i.e., temporal information). 
This observation motivates us to use the spatial-temporal information to accurately divide video into chunks, which exhibits a different design space to overfit video data. With the different levels of information density within each patches, the key insight is to cluster patches that has similar texture complexity across all frames, and use one SR model to overfit each patch group. In this way, the number of video chunks and their associated SR models are effectively reduced, which improves the encoding efficiency regardless the duration of videos. Meanwhile, a compact SR model can be adopted without causing quality degradation because each SR model only overfits one specific video content with similar texture complexity. Additionally, when the spatial-temporal data is properly scheduled, our method can be extended to a joint training manner which generates a single SR model for the entire video.



\subsection{Spatial-Temporal Data Overfitting}\label{sec:method}

In this section, we introduce a novel spatial-temporal data overfitting approach, STDO, which efficiently encodes HR videos into LR videos and their corresponding SR models, and achieves outperforming super-resolution quality \& speed at user end. 

Suppose the video time length is $T$. General method to train an SR model would firstly divide the video into frames, and slice each frame into multiple non-overlapping image patches. All patches across all dimensions such as their locations in the frame or time compose a complete video. For a given video with the dimension $W \times H$, and the desired patch size $w \times h$, the patch is denoted as $P_{i,j,t}$, where $i \in [0,I)$, $j \in [0,J)$, and $t \in [0,T)$. Note that $I=\lfloor \frac{W}{w} \rfloor$ and $J=\lfloor \frac{H}{h} \rfloor$ are integer numbers, then the training set for this specific video is $\mathcal D=\{P_{n}\}_{0}^{N}$ where $N=I\times J\times T$ is the total number of patches.

Note that $\mathcal D$ contains all patches across all dimensions. We use a pretrained SR model $f_{0}(\cdot)$ to super-resolve all of the LR patches and compute their PSNRs with the HR patches. As illustrated in Figure~\ref{fig:intro_1}, we find that the distribution of the PSNR is usually not uniform, and shows a clear boundary regarding the content of the video. We divide the training set $\mathcal D$ into multiple chunks by grouping patches with similar PSNRs, and form a set of chunks as $\Omega=\{ \hat{\mathcal D}_{0}, \hat{\mathcal D}_{1}, \dots, \hat{\mathcal D}_{k} \}$, in which
\begin{equation}
\hat{\mathcal D}_{k} = \{P_{n}| P_{n} \in {\mathcal D}, PSNR(f_{0}(P_{n})) \in [\lambda_{k1}, \lambda_{k2}) \},
\end{equation}
where $\lambda$ is the threshold. We set the first chunk $\hat{\mathcal D}_{0}$ to be group of the patches with lowest PSNRs, and we list all chunks in ascending order, which means $\hat{\mathcal D}_{k}$ to be the patches with the highest PSNRs. In this way, we separate training data by their spatial-temporal information in a one-shot manner, which is usually done in seconds and can be considered negligible compared to the training process. In this paper, we empirically divide $\mathcal D$ evenly. 
Finally, we train an SR model $f_{sr_{k}}(\mathbf w_k; \hat{\mathcal D}_{k})$ to overfit each video chunk $\hat{\mathcal D}_{k}$.
Experimental results indicate better performance on both video quality and execution speed. Our empirical analysis is that by accurately identifying and grouping the data with similar information density (i.e, texture complexity) into chunks, each SR model becomes easier to ``memorize" similar data in an overfitting task, and subsequently demands smaller SR models that can be executed in a real-time fashion.

\begin{table}[t]
\small
\centering
\caption{Video super-resolution results comparison of vanilla STDO training and using \emph{only one} model from STDO trained with the most informative chunk $\hat{\mathcal D}_{0}$ and least informative chunk $\hat{\mathcal D}_{k}$, respectively. We also include the video super-resolution results with one model trained with all data~\cite{yeo2017will}.}
\vspace{-0.3cm}
\scalebox{0.8}{
\begin{tabular}{ll|ccc|ccc}
\toprule & \bf {} & \multicolumn{3}{c|}{\bf{vlog-15s}} & \multicolumn{3}{c}{\bf {vlog-45s}}  \\
\bf{Model} & \bf{Method} & $\times$2 & $\times$3 & $\times$4  & $\times$2 & $\times$3 & $\times$4 \\
\midrule 
\multirow{4}{*}{WDSR} & awDNN~\cite{yeo2017will} & 49.24 & 45.30 & 43.33 & 48.02 & 44.16 & 42.19 \\
& STDO & 50.58 & 46.43 & 44.62 & 49.76 & 45.95 & 43.99 \\
& STDO$|_{\hat{\mathcal D}_{0}}$ & 50.42 & 45.99 & 44.18 & 49.51 & 45.63 & 43.75 \\
& STDO$|_{\hat{\mathcal D}_{k}}$ & 46.89 & 42.63 & 40.25 & 44.89 & 41.07 & 38.87 \\
\bottomrule
\end{tabular}}
\label{tab:table2}
\vspace{-0.2cm}
\end{table}

\begin{table*}[t]
\linespread{0.9}
\small
\centering
\caption{Comparison results of STDO with different data overfitting methods on different SR backbones.}
\vspace{-0.5em}
\scalebox{0.85}{
\begin{tabular}{llp{1.3cm}p{1.3cm}p{1.3cm}p{1.3cm}p{1.3cm}p{1.3cm}p{1.3cm}p{1.3cm}p{1.3cm}}
\toprule
& \bf { Data } & \multicolumn{3}{c}{\bf { game-15s }}& \multicolumn{3}{c}{\bf { inter-15s }} & \multicolumn{3}{c}{\bf {vlog-15s}}  \\
\bf { Model } & \bf { Scale } & $\times$2 & $\times$3 & $\times$4 & $\times$2 & $\times$3 & $\times$4 & $\times$2 & $\times$3 & $\times$4 \\
\midrule 
\multirow{4}{*}{ESPCN} & awDNN~\cite{yeo2017will} & 37.94 & 32.85 & 29.97 & 40.43 & 35.36 & 29.91 & 46.41 & 42.90 &39.65 \\
& NAS~\cite{yeo2018neural} & 37.58 & 32.71 & 30.59 & 40.62 & 35.42 & 30.43 & 46.53 & 43.01 & 39.98 \\
& \text {CaFM\cite{liu2021overfitting} } & 38.07 & 33.14 & 30.96 & 40.71 & 35.54 & 30.47 & 47.02 & 43.20 & 40.16 \\
& \bf{STDO} & \textbf{38.61} & \textbf{33.57} & \textbf{31.30}  & \textbf{42.65} & \textbf{35.63} & \textbf{30.63} & \textbf{47.11}  & \textbf{43.25} & \textbf{40.73} \\
\midrule
\multirow{4}{*}{SRCNN} & awDNN~\cite{yeo2017will} & 36.08 & 31.94 & 29.90 & 40.46 & 33.95 & 28.78 & 46.69 & 42.41 &39.71 \\
& NAS~\cite{yeo2018neural} & 36.27 & 32.08 & 29.94 & 40.70 & 34.01 & 28.84 & 46.78 & 42.53 & 39.76 \\ 
& CaFM\cite{liu2021overfitting} & 36.63 & 32.21 & 29.98 & 40.76 & 34.08 & 29.93 & 46.98 & 42.62 &39.81 \\
& \bf{STDO} & \textbf{37.59} & \textbf{32.67} & \textbf{30.64} & \textbf{42.28} & \textbf{34.26} & \textbf{30.05} &\textbf{47.06}  & \textbf{42.78} & \textbf{39.90}\\
\midrule 
\multirow{4}{*}{VDSR}& awDNN~\cite{yeo2017will} & 41.27 & 35.03 & 32.16 & 44.16 & 35.99 & 30.65 & 48.18 & 43.03 & 41.07\\
& NAS~\cite{yeo2018neural} & 42.53 & 35.97 & 33.86 &  44.71  & 36.57 & 31.05 & 48.49 &43.41 &41.33\\
& CaFM\cite{liu2021overfitting} & 43.02 & 36.17 & 33.98 & 44.85 & 36.46 & 31.08 & 48.61 & 43.62 &41.49\\
& \bf{STDO} & \textbf{43.56} & \textbf{36.71} & \textbf{35.02} & \textbf{45.16} & \textbf{36.81} &\textbf{33.43}  &\textbf{48.75}  & \textbf{43.82} & \textbf{41.71} \\
\midrule
\multirow{4}{*}{EDSR} &   awDNN~\cite{yeo2017will} & 42.24 & 35.88 & 33.44 & 43.06 & 37.89 & 34.94 & 48.87 & 44.51 & 42.58\\ 
& NAS~\cite{yeo2018neural} & 42.82 & 36.42 & 34.00 & 45.06 & 38.38 & 35.47 & 49.10 & 44.80 & 42.83 \\
&CaFM~\cite{liu2021overfitting} & 43.13 & 37.04 & 34.47 & 45.35 & 38.66 & 35.70 & 49.30 & 45.03 & 43.12  \\
& \bf{STDO} & \textbf{44.93} & \textbf{37.80} & \textbf{35.47}  & \textbf{45.91}  & \textbf{39.26} & \textbf{36.76} & \textbf{50.24} & \textbf{45.68}  & \textbf{43.46} \\ 
\midrule
\multirow{4}{*}{WDSR}&  awDNN~\cite{yeo2017will}&43.36 & 37.12 & 34.62 & 44.83 & 39.05 & 35.23 & 49.24 & 45.30 & 43.33  \\ 
& NAS~\cite{yeo2018neural} &44.17  & 38.23 & 36.02 & 45.43 & 39.71 & 36.54 & 49.98 & 45.63 & 43.51 \\
& CaFM\cite{liu2021overfitting} & 44.23 & 38.55 &36.30  &45.71  & 39.92  & 36.87 & 50.12 & 45.87 & 43.79\\
& \bf{STDO} & \textbf{45.75} &  \textbf{40.17} & \textbf{38.62} & \textbf{46.34} & \textbf{41.13} & \textbf{38.76} & \textbf{50.58} & \textbf{46.43} & \textbf{44.62} \\ 

\toprule

&  & \multicolumn{3}{c}{\bf { game-45s }}& \multicolumn{3}{c}{\bf { inter-45s }} & \multicolumn{3}{c}{\bf {vlog-45s}}  \\
&  & $\times$2 & $\times$3 & $\times$4 & $\times$2 & $\times$3 & $\times$4 & $\times$2 & $\times$3 & $\times$4 \\
\midrule 
\multirow{4}{*}{ESPCN} & awDNN~\cite{yeo2017will} & 35.42 & 30.63 & 28.65 & 38.64 & 31.97 & 28.32 & 45.71 & 41.40 & 39.20 \\
& NAS~\cite{yeo2018neural} & 35.55 & 30.67 & 28.74 & 38.81 & 32.14 & 28.61 & 45.81 & 41.52 & 39.29 \\
& CaFM\cite{liu2021overfitting} & 36.09 & 31.06 & 29.05 & 38.88 & 32.22 & 28.75 & 46.19 & 41.72 & 39.52  \\
& \bf{Ours} & \textbf{37.75} & \textbf{32.29} & \textbf{29.96} & \textbf{41.20} & \textbf{32.48} & \textbf{29.09} & \textbf{46.33} & \textbf{42.26} & \textbf{40.26}  \\
\midrule
\multirow{4}{*}{SRCNN} & awDNN~\cite{yeo2017will} & 35.05 & 30.50 & 28.59 & 38.66 & 31.78 & 28.25 & 45.87 & 41.58 &39.29 \\
& NAS~\cite{yeo2018neural} & 35.15 & 30.55 & 28.61 & 38.79 & 31.93 & 28.38 & 45.95 & 41.66 & 39.36 \\ 
& CaFM\cite{liu2021overfitting} & 35.49 & 30.63 & 28.66 & 38.88 & 32.02 & 28.48 & 46.18 & \textbf{41.85} & 39.52 \\
& \bf{STDO} & \textbf{36.74} & \textbf{31.46} & \textbf{29.37} & \textbf{41.15} & \textbf{32.17} & \textbf{28.65} & \textbf{46.33} & 41.81 & \textbf{39.69} \\
\midrule 
\multirow{4}{*}{VDSR}& awDNN~\cite{yeo2017will } & 40.29 & 34.53 & 31.28 & 41.99 & 33.80 & 30.34 & 47.61 & 42.92 & 40.94 \\
& NAS~\cite{yeo2018neural} & 41.37 & 34.92 & 32.42 & 42.40 & 34.53 & 31.10 & 47.88 & 43.33 & 41.23\\
& CaFM\cite{liu2021overfitting}  & 41.92 & 35.56 & 33.16 & 42.86 & 34.49 & 30.95 & 48.00 & 43.50 & 41.38\\
& \bf{STDO} &\textbf{42.65}  & \textbf{36.23}  &\textbf{33.76} &\textbf{43.36}& \textbf{35.64} & \textbf{31.77} & \textbf{48.17} & \textbf{43.67} & \textbf{41.49}  \\
\midrule
\multirow{4}{*}{EDSR}& awDNN~\cite{yeo2017will}  & 42.11 & 35.75 & 33.33 & 42.73 & 34.49 & 31.34 & 47.98 & 43.58 & 41.53\\ 
& NAS~\cite{yeo2018neural} & 43.22 & 36.72 & 34.32 & 43.31 & 35.80 & 32.67 & 48.48 & 44.12 & 42.12 \\
& CaFM\cite{liu2021overfitting} & 43.32 & 37.19 & 34.61 & 43.37 & 35.62 & 32.35 & 48.45 & 44.11 & 42.16\\
& \bf{STDO} & \textbf{45.65} & \textbf{39.93} & \textbf{37.24} & \textbf{44.52} & \textbf{38.28} & \textbf{35.51} & \textbf{49.84} & \textbf{45.47} & \textbf{43.07} \\ 
\midrule
\multirow{4}{*}{WDSR}&  awDNN~\cite{yeo2017will} & 42.61 & 36.17 & 33.85 & 42.94 & 34.71 & 31.81 & 48.02 & 44.16 & 42.19\\ 
& NAS~\cite{yeo2018neural} & 43.72 & 37.25 & 34.93 & 43.41 & 36.05 & 33.11 & 48.52 & 44.75 & 42.80 \\
& CaFM\cite{liu2021overfitting} & 43.97 & 37.64 & 35.12 & 43.52 & 36.03 & 32.97 & 48.51 & 44.72 & 42.87\\
& \bf{STDO} & \textbf{45.71} & \textbf{40.33} & \textbf{37.76} & \textbf{44.54} & \textbf{38.72} & \textbf{36.03} & \textbf{49.76} & \textbf{45.95} & \textbf{43.99} \\ 
\bottomrule
\end{tabular}}
\label{tab:different_backbones}
\vspace{-0.4cm}
\end{table*}

\subsection{Data-Aware Joint Training}

In Section~\ref{sec:method}, our method significantly reduces the number of video chunks and overfitting SR models by effectively utilizing spatial-temporal information of each patch in the video. In this section, we extend our method by generating a single SR model for the entire video, which further reduces the storage requirement with negligible quality drop. From the set of chunks $\Omega \in \mathbb{R}^{k}$ and all SR models, we demonstrate PSNR in Table~\ref{tab:table2} by using \emph{only one} SR model to super-resolve the entire video. Somehow surprisingly, we find out that using the model trained with $\hat{\mathcal D}_{0}$ (i.e., the most informative chunk) experiences a moderate quality drop, and achieves similar or higher PSNR compared to the model trained with all patches. Meanwhile, the model trained with $\hat{\mathcal D}_{k}$ has a severe quality degradation. We argue that low PSNR patches usually contain rich features, and contribute more to the model learning, which improves the generalization ability to the less informative data~\cite{toneva2018empirical,yuan2021mest}.
Motivated by the above observation, we propose a joint training technique, which carefully schedules the spatial-temporal data participated in training to train a single SR model that overfits the entire video. Concretely, we keep all patches for $\hat{\mathcal D}_{0}$, and remove the entire $\hat{\mathcal D}_{k}$. For the rest of the chunks, we randomly sample a portion of the patches from each chunk, while gradually decreasing the proportion of the data sampled. We train a single model by solving the following optimization problem using the joint dataset
\begin{align}\label{eq: prob}
    \begin{array}{cl}
\displaystyle \minimize_{\mathbf w}  & f_{joint}( \mathbf w; \mathcal D_{joint}) \\
     \st     &  \mathcal D_{joint} \in \{ \hat{\mathcal D}_{0}, \rho_1 \odot\hat{\mathcal D}_{1}, \dots, \rho_{k-1}\odot\hat{\mathcal D}_{k-1} \}, \\
     & \sum_{i=0}^{k-1} \| \rho_{i}\odot\hat{\mathcal D}_{i} \| = \mu,
    \end{array}
\end{align}
where $\odot$ is the sampling operation with pre-defined proportion $\rho$, and $\mu$ is a hyper-parameter that control the size of the joint dataset.

\section{Experimental Results}

In this section, we conduct extensive experiments to prove the advantages of our proposed methods. To show the effects of our methods, we apply our proposed STDO and JSTDO to videos with different scenes and different time lengths. The detailed information on video datasets and implementations are shown in Section~\ref{sec:preli}. In Section~\ref{sec:1}, we compared our method with time-divided method using different videos and different SR models, which show that STDO achieves outperforming video super-resolution quality as well as using lowest computation budgets. In section~\ref{sec:2}, we demonstrate the results of our single SR model obtained by JSTDO and show that JSTDO effectively exploits heterogeneous information density among different video chunks to achieve better training performance. In Section~\ref{sec:mobile_speed}, we deploy our model on an off-the-shelf mobile phone to show our model can achieve real-time video super-resolution. In Section~\ref{sec:ablation}, we show our ablation study on key parameters used in STDO and JSTDO methods, such as the different number of chunks, training data scheduling, and long video with multiple scene conversions.

\begin{table}[t]
\small
\centering
\caption{Computation cost for different backbones with VSD4K video game-45s. We include the computation cost for the models with different resolution upscaling factors.}
\scalebox{0.99}{
\begin{tabular}{lcccc}
\toprule 
\bf { Model } & \bf { Scale } &  \bf{FLOPs} &\textbf{CaFM~\cite{liu2021overfitting}} &\textbf{STDO}  \\
\midrule \multirow{3}{*}{\text { ESPCN }} & $\times$2 & 0.14G & 36.09 & 37.75   \\
&$\times$3 & 0.15G & 31.06 & 32.29  \\
&$\times$4  & 0.16G & 29.05 & 29.96   \\
\midrule \multirow{3}{*}{\text { SRCNN }} & $\times$2  & 0.64G &  35.49 & 36.74   \\
&$\times$3 & 1.45G & 30.63 & 31.46  \\
&$\times$4  & 2.58G & 28.66 & 29.37   \\
\midrule \multirow{3}{*}{\text { VDSR }} & $\times$2  & 6.15G & 41.92 & 42.65   \\
& $\times$3 & 13.85G & 35.56 & 36.23  \\
&  $\times$4  & 20.62G & 33.16 & 33.76   \\
\midrule \multirow{3}{*}{\text { EDSR }} & $\times$2  & 3.16G & 43.32 & 45.65   \\
&$\times$3 & 3.60G & 37.19 & 39.93 \\
&$\times$4  & 4.57G & 34.61 &  37.24  \\
\midrule \multirow{3}{*}{\text { WDSR }} & $\times$2  & 2.73G & 43.97 & 45.71   \\
&$\times$3 & 2.74G & 37.64 &40.33  \\
&$\times$4  & 2.76G & 35.12 & 37.76   \\
\bottomrule
\end{tabular}}
\label{tab:CaFM_compare}
\end{table}

\subsection{Datasets and Implementation Details}\label{sec:preli}

In the previous video super-resolution works, most video datasets ~\cite{nah2019ntire, xue2019video} for super-resolution only provide several adjacent frames as a mini-video. Those mini-video sets are not suitable for a network to overfit. Therefore, we adopt the VSD4K collected in~\cite{liu2021overfitting}. In this video dataset, there are 6 video categories including: vlog, game, interview, city, sports, dance. Each of the category contains various video lengths. We set the resolution for HR videos to 1080p, and LR videos are generated by bicubic interpolation to match different scaling factors.

We apply our approach to several popular SR models including ESPCN~\cite{shi2016real}, SRCNN~\cite{dong2015image}, VDSR~\cite{kim2016accurate}, EDSR~\cite{lim2017enhanced}, and WDSR~\cite{yu2018wide}. 
During training, we use patch sizes of 48$\times$54 for scaling factor 2 and 4, and 60$\times$64 for scaling factor 3 to accommodate HR video size, and the threshold value $\lambda$ is set to split the patches evenly into chunks.
Regarding the hyperparameter configuration of training the SR models, we follow the setting of ~\cite{lim2017enhanced,yu2018wide,liu2021overfitting}. We adopt Adam optimizer with $\beta_1 = 0.9$, $\beta_2 = 0.009$, $\epsilon=10^{-8}$ and we use L1 loss as loss function. For learning rate, we set $10^{-3}$ for WDSR and $10^{-4}$ for other models with decay at different training epochs. We conduct our experiment on EDSR model with 16 resblocks and WDSR model with 16 resblocks.

\begin{figure}[t]
\centering
\includegraphics[width=1\columnwidth]{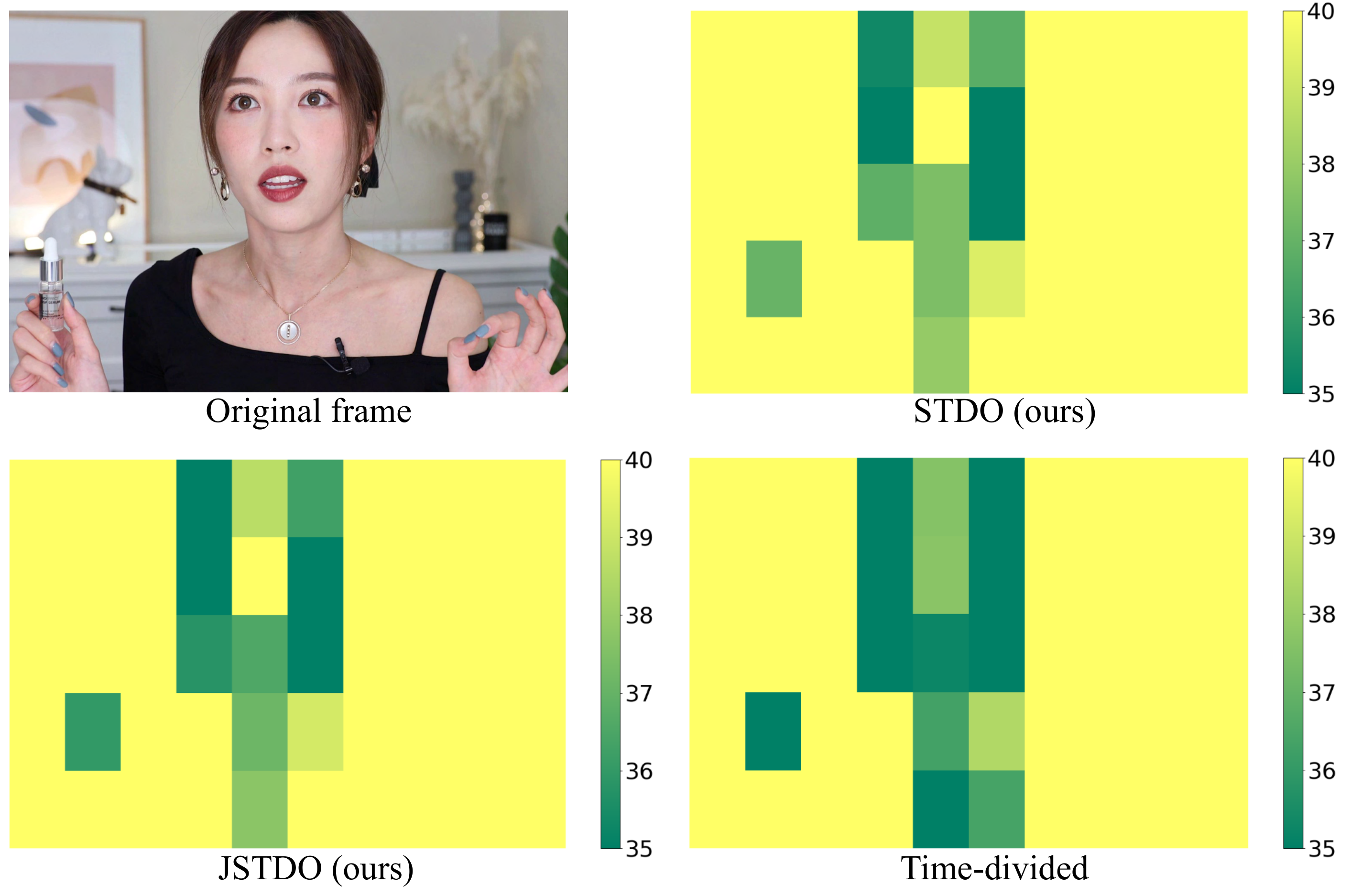}
\caption{PSNR heatmaps of super-resolving an LR video with different methods. STDO and JSTDO has similar value in the key content zone (i.e., body), and outperform time-divided method.}
\label{fig:join_show}
\end{figure}

\subsection{Compare with the State-of-the-Art Methods}\label{sec:1}

\begin{figure*}[t]
\centering
\includegraphics[width=0.95\textwidth]{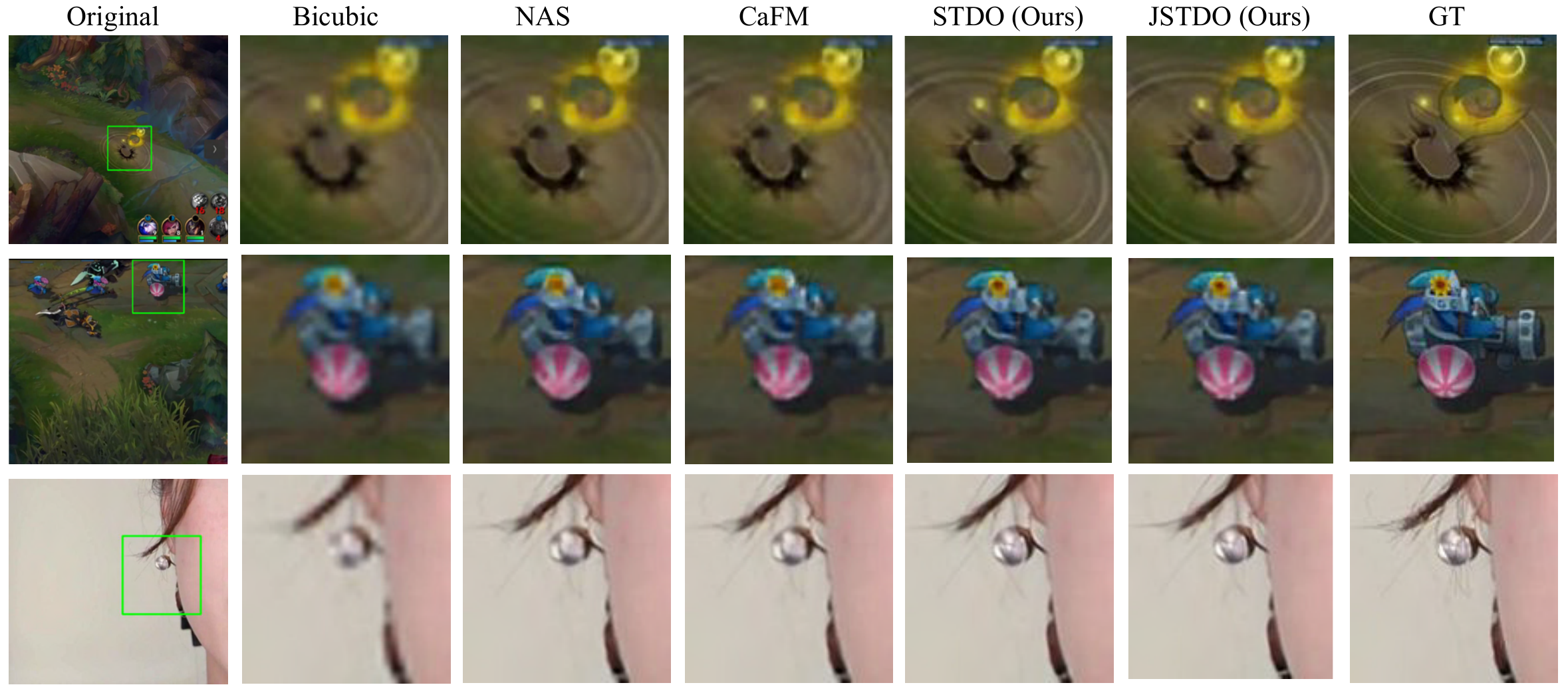}
\caption{Super-resolution quality comparison with random video frame using STDO and JSTDO with baseline methods.}
\label{fig:sr_illustration}
\vspace{-0.4cm}
\end{figure*}

In this section, we compare our method with the state-of-the-art (SOTA) that either use general model overfitting or time-divided model overfitting on different SR backbones. Due to space limits, we sample three video categories from VSD4K, and test on two different video time lengths as 15$s$ and 45$s$. Our results are shown in Table~\ref{tab:different_backbones}. We compare with the state-of-the-art neural network-based SR video delivery methods, such as awDNN~\cite{yeo2017will} where a video is overfitted by a model, NAS~\cite{yeo2018neural} that splits a video into multiple chunks in advance and overfit each of the time-divided chunk with independent SR model, and CaFM~\cite{liu2021overfitting} that uses time-divided video chunk and single SR model with hand-crafted module to overfit videos. 
For our implementation of STDO, we divide the spatial-temporal data into 2, 4, and 6 chunks respectively, and report the best results. We adjust batch size while training to keep the same computation cost, and we show the comparison by computing the PSNR of each method. It can be seen that our method can exceed the SOTA works consistently on different backbones. 

With STDO, each SR model is only overfitting one video chunk that has similar information density, which makes it suitable to use smaller and low capacity SR models that has low computations.
In Table~\ref{tab:CaFM_compare}, we demonstrate the computation cost on each model. 
From the results, we notice that with the relatively new model such as VDSR, EDSR, and WDSR, when the computation drops below 3 GFLOPs, time-divided method experiences significant quality degradation, while STDO maintains its performance or even achieves quality improvements. When using extremely small networks such as ESPCN or SRCNN, time-divided methods PSNR drops quickly, while STDO still achieves 0.7 $\sim$ 1.7 dB better performance.

\begin{table}[t]
\centering
\caption{Comparison between STDO and JSTDO regarding PSNR and total number of model parameters with game-15s from VSD4K. We compute the PSNR difference of the two methods.}
\scalebox{0.83}{
\begin{tabular}{llcccc}
\toprule
\textbf{Model}& \textbf{Method} & \textbf{\#Chunks}& \textbf{\#Models}& \textbf{\#Param.} & \textbf{PSNR} \\ \midrule
\multirow{3}{*}{WDSR$\times 2$} & STDO & 4 & 4 &   4.8M        & 45.75 \\
& {JSTDO}  &4       & \textbf{1} & \textbf{1.2M} & 45.46 \\ \cline{2-6} \\[-1em]
& \multicolumn{5}{c}{$\Delta_{PSNR}$: 0.29} \\
\midrule
\multirow{3}{*}{WDSR$\times 3$} & STDO & 4 & 4 & 4.8M         & 40.17 \\
& {JSTDO}  &4       & \textbf{1} & \textbf{1.2M} & 39.87 \\ \cline{2-6} \\[-1em]
& \multicolumn{5}{c}{$\Delta_{PSNR}$: 0.30} \\
\midrule
\multirow{3}{*}{WDSR$\times 4$} & STDO & 4 & 4 &  4.8M         & 38.62 \\
& {JSTDO}  &4      & \textbf{1} & \textbf{1.2M} & 38.14 \\ \cline{2-6} \\[-1em]
& \multicolumn{5}{c}{$\Delta_{PSNR}$: 0.48} \\
\bottomrule
\end{tabular}
}
\label{tab:joint_training}
\end{table}


\begin{figure}[b]
\centering
\vspace{-0.6cm}
\includegraphics[width=1\columnwidth]{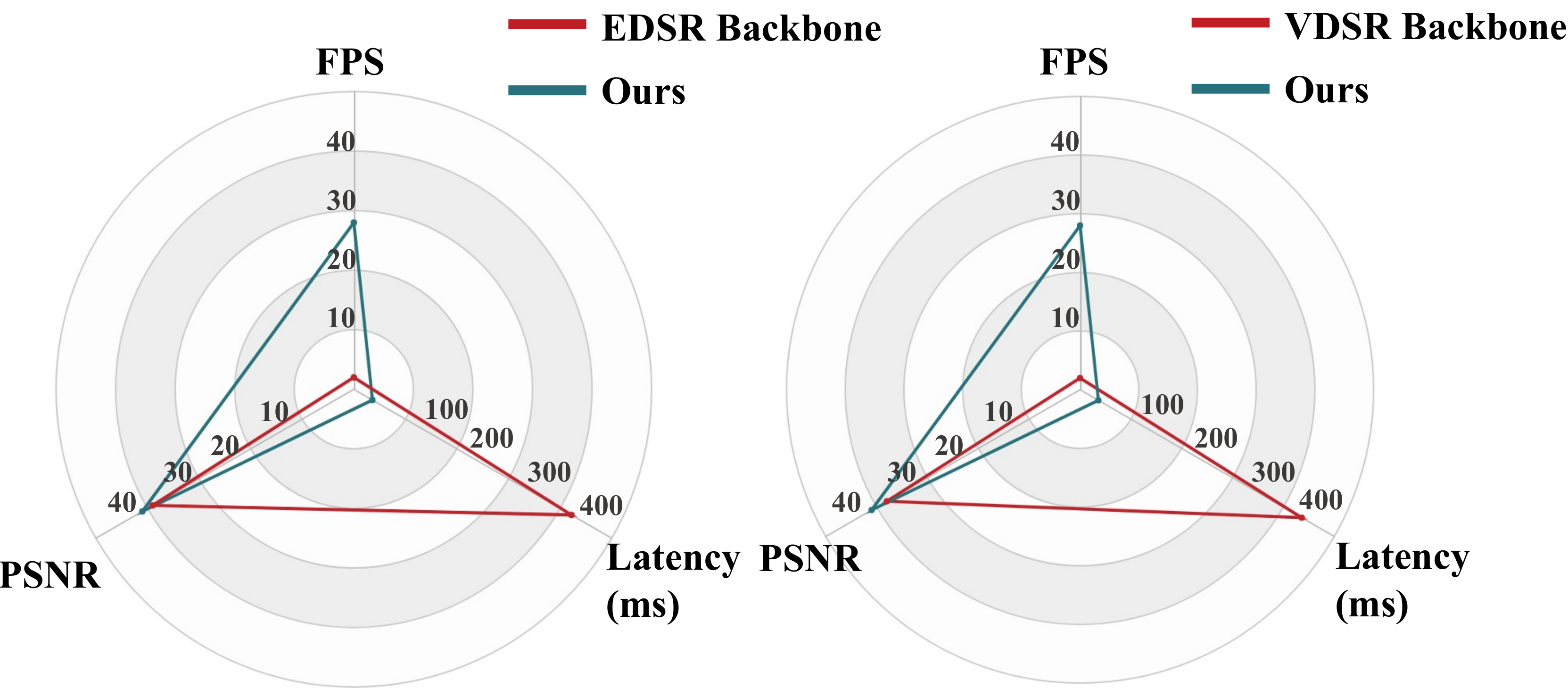}
\caption{Execution speed and video quality comparison between STDO and \cite{liu2021overfitting}\cite{yeo2017will} respectively using an Samsung mobile phone.}
\label{fig:radar_edsr}
\end{figure}

\begin{figure*}[t]
\centering
\includegraphics[width=0.95\textwidth]{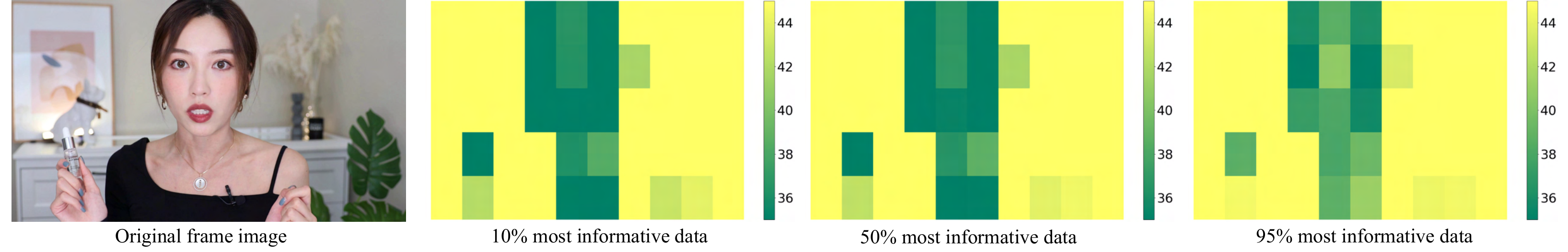}
\caption{PSNR heatmap with different ratio of $\hat{\mathcal D}_{0}$ in JSTDO. More informative data participates in joint training achieves better PSNR.}
\label{fig:ab_heat}
\end{figure*}

\subsection{Data-Aware Joint Training}\label{sec:2}

In this part, we show the results of reducing the number of SR models to a single model by data-aware joint training with the spatial-temporal data overfitting (JSTDO) in Table~\ref{tab:joint_training}. We conduct our experiment on the relatively latest model WDSR~\cite{yu2018wide} to chase a better recovering performance. In our experiments, we use 4 chunks for WDSR implementation, and compare the total number of parameters and PSNR of STDO with those of JSTDO which only uses \emph{one} SR model with the same model architecture used by STDO. From the results, we can clearly see that JSTDO effectively reduces the overall model size from 4.8 MB to 1.2 MB, while the PSNR of JSTDO has only negligible degradation compared with STDO. Please note that even with some minor quality degradation, JSTDO still outperforms baseline methods in both super-resolution quality and total parameter counts.

We also plot the PSNR heatmaps when using STDO and JSTDO for video super-resolution, and compare with the traditional time-divided method. As showing in Figure~\ref{fig:join_show}, we randomly select one frame in a vlog-15s video that is super-resolved  from 270p to 1080p ($\times$4) by CaFM~\cite{liu2021overfitting}, STDO, and JSTDO. The heatmaps clearly demonstrate that our methods achieve better PSNR in the key content zone in the frame. Meanwhile, another key observation can be drawn: the JSTDO heatmap has similar patterns with the one using STDO, which further proves that the joint training technique using carefully scheduled spatial-temporal data effectively captures the key features, while not losing the expressive power towards the low information density data. We also show the qualitative comparison in Figure~\ref{fig:sr_illustration}.

\subsection{Deployment on Mobile Devices}\label{sec:mobile_speed}

One of the many benefits by using STDO is that we can use smaller (i.e., low model capacity \& complexity) SR models to perform data overfitting. The reason is that the patches in each chunk are relatively similar, especially for some short videos, which makes it easier for smaller models to ``memorize" them. Subsequently, unlike CaFM~\cite{liu2021overfitting}, no handcrafted modules are needed for both STDO and JSTDO methods, which further reduces the compilation burden on the end-user devices.

We deploy the video chunks alongside with the overfitting models of STDO on a Samsung Galaxy S21 with Snapdragon 888 to test execution performance. Each patch will have a unique index to help assemble into frames. Our results are shown in Figure~\ref{fig:radar_edsr}. We set the criteria for real-time as latency less than 50 $ms$ and FPS greater than 20 on the mobile devices according to~\cite{zhan2021achieving}. The result shows that our method achieves 28 FPS when super-resolving videos from 270p to 1080p by WDSR, and it is significantly faster in speed and better in quality than other models such as EDSR or VDSR that are originally used in other baseline methods ~\cite{yeo2018neural,yeo2017will,liu2021overfitting,yeo2020nemo}. Please note that the capability of using small scale SR models to accelerate execution speed is ensured by the high super-resolution quality achieved by spatial-temporal data overfitting method.

\subsection{Ablation Study}\label{sec:ablation}

\begin{figure}[h]
\centering
    \begin{minipage}[b]{1.0\textwidth}
        \begin{subfigure}[b]{0.26\textwidth}
             \centering
             \includegraphics[width=\textwidth]{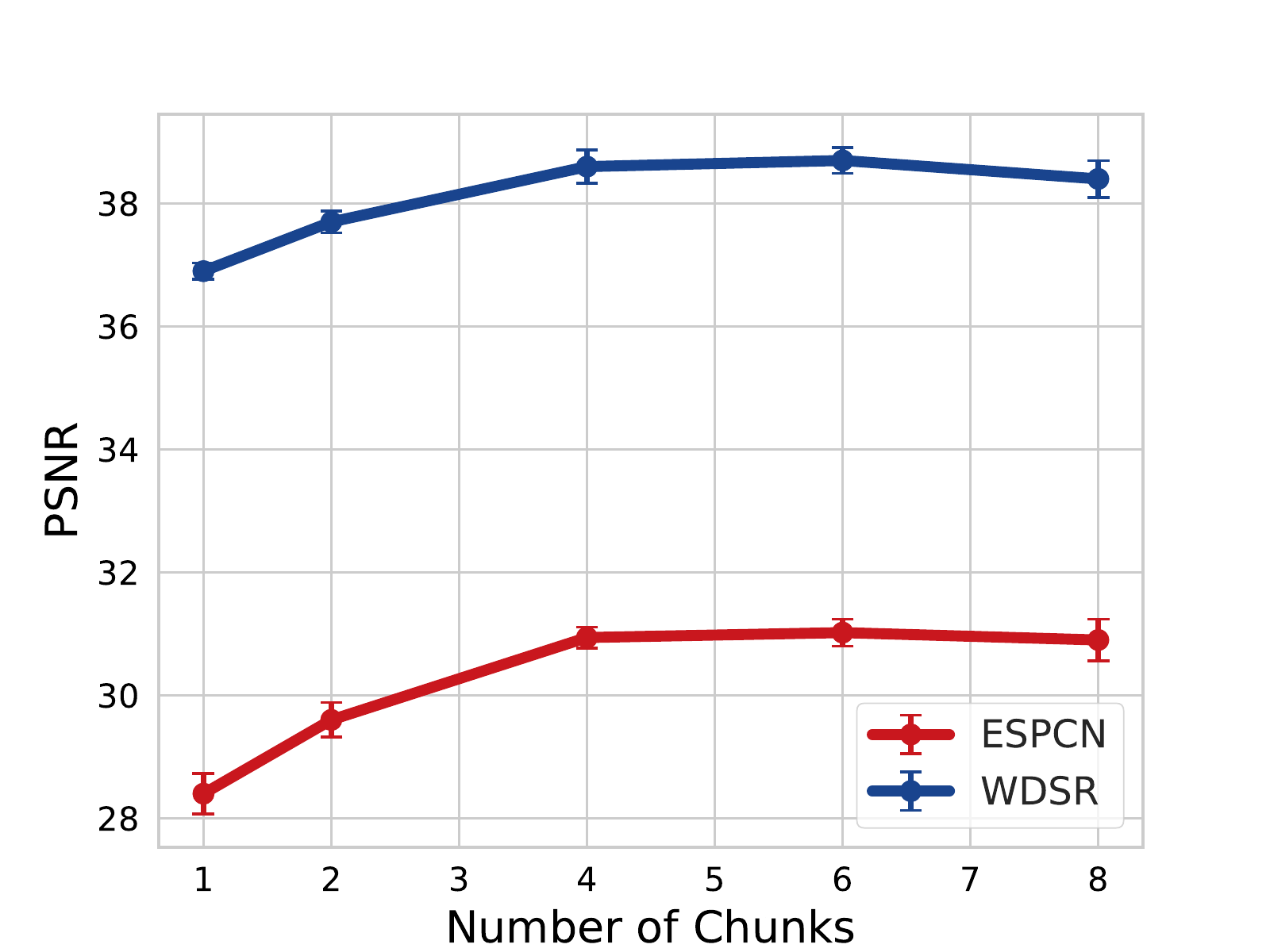}
             \caption{}
             \label{fig:ab1}
        \end{subfigure}\hspace{-1em}
        \begin{subfigure}[b]{0.26\textwidth}
             \centering
             \includegraphics[width=\textwidth]{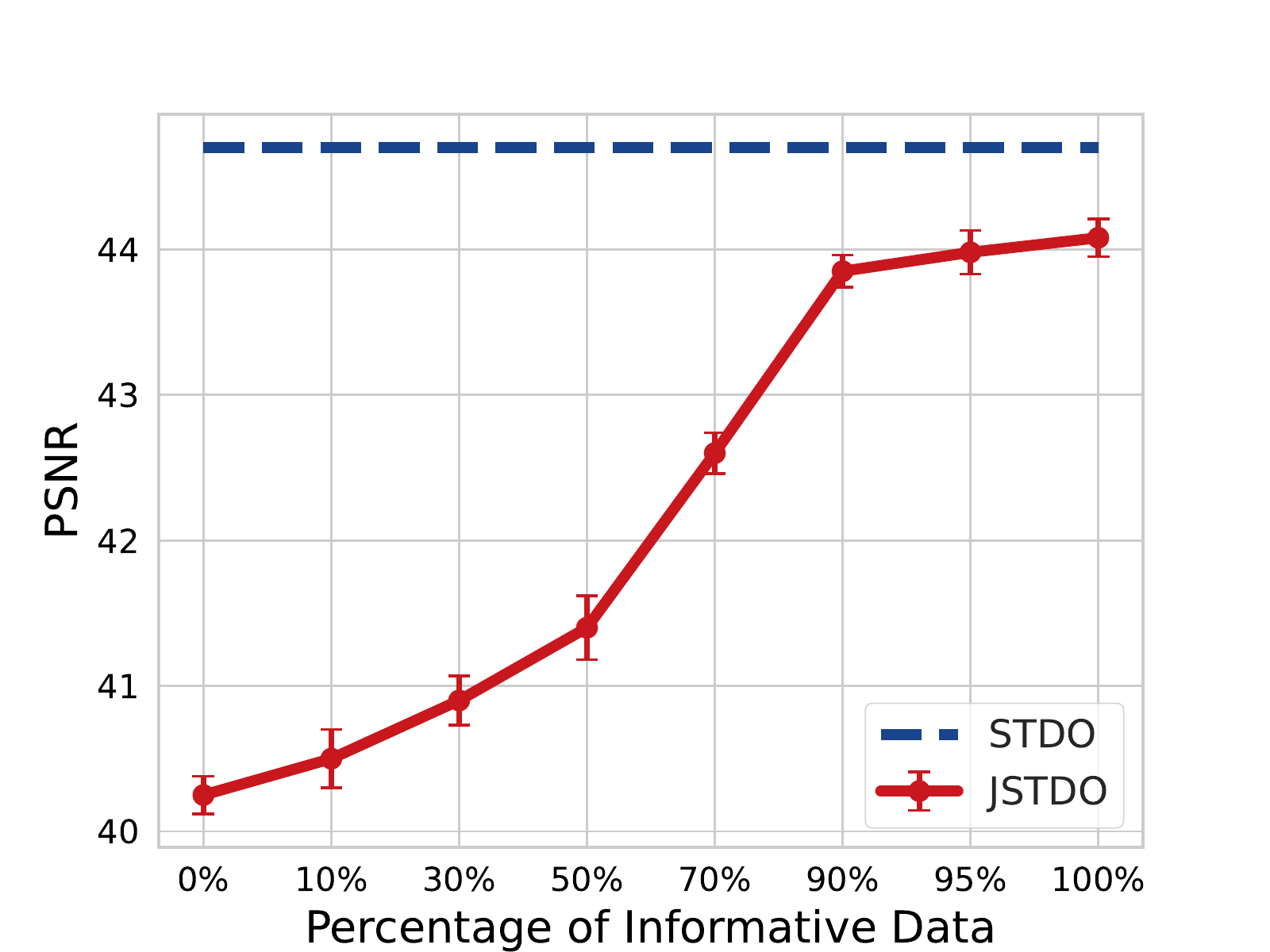}
             \caption{}
             \label{fig:ab2}
        \end{subfigure}
    \end{minipage}
\caption{(a) PSNR of video game-15s on different number of chunks, (b) Comparison of different data schedule in joint training.}
\vspace{-1.0em}
\label{fig:abation}
\end{figure}

\noindent\textbf{$\rhd$ Different number of chunks in STDO.}
Previous experiments have proved that STDO brings performance improvement when we take account of the spatial-temporal information. In this ablation study, we vary the number of chunks and evaluate their video super-resolution quality. We set the number of chunks with the range of 1 (i.e., single model overfitting) to 8, and we plot the PSNR trends using ESPCN and WDSR in Figure~\ref{fig:ab1}. We observe that ESPCN and WDSR demonstrate similar trends when the number of chunks increases, and better results can be obtained when we divide video into $\sim$4 chunks, which consolidates our claim that STDO uses fewer chunks compared to time-divided methods.


\noindent\textbf{$\rhd$ Data scheduling in joint training.}
In JSTDO, we vary the sampling proportion by increasing patches from $\hat{\mathcal D}_{0}$ while decreasing the proportion of patches in $\hat{\mathcal D}_{k}$, and adjusting sampling proportion for other chunks accordingly to maintain the same amount of training patches. The evaluation results are shown in Figure~\ref{fig:ab2}, where we observe that when more informative data participates in training, the overall video super-resolution quality increases. Same patterns can be seen in Figure~\ref{fig:ab_heat}, where the heatmaps show relatively high PSNR in the key content zone when the SR model is trained with more informative data.

\noindent\textbf{$\rhd$ Long video with multiple scene conversions.} 
We combine the game-45$s$ video and the vlog-45$s$ video together into a 90$s$ long video which contains multiple scene conversions. The STDO result of the long video trained on WDSR is 39.92 dB with scaling factor 4, which is close to the average value of overfitting two videos (43.99 dB and 37.76 dB).
Therefore, it can be proved that our design can still maintain good performance for long videos where multiple scene conversions exist.

\section{Conclusion}
In this paper, we introduce a novel spatial-temporal data overfitting approach towards high-quality and efficient video resolution upscaling tasks at the user end. We leverage the spatial-temporal information based on the content of the video to accurately divide video into chunks, then overfit each video chunk with an independent SR model or use a novel joint training technique to produce a single SR model that overfits all video chunks. We successfully keep the number of chunks and the corresponding SR models to a minimum, as well as obtaining high super-resolution quality with low capacity SR models. We deploy our method on the mobile devices from the end-user and achieve real-time video super-resolution performance.

\section{Acknowledgment}
This work is partly supported by the National Science Foundation IIS-1949640, CNS-2008049, CNS-2232048, and CNS-2204445, and Air Force Office of Scientific Research FA9550-20-1-0090. Any opinions, findings, and conclusions or recommendations expressed in this material are those of the authors and do not necessarily reflect the views of NSF and AFOSR.

{\small
\bibliographystyle{ieee_fullname}
\bibliography{egbib}
}

\end{document}